# Manipulation of Camera Sensor Data via Fault Injection for Anomaly Detection Studies in Verification and Validation Activities For AI

Alim Kerem Erdoğmuş[1]

kerem.erdogmus@inovasyonmuhendislik.com

Mustafa Karaca[1]

mustafa.karaca@inovasyonmuhendislik.com

Assist. Prof. Dr. Ugur Yayan[2]

ugur.yayan@ogu.com.tr

January 14, 2022

## Abstract

In this study, the creation of a database consisting of images obtained as a result of deformation in the images recorded by these cameras by injecting faults into the robot camera nodes and alternative uses of this database are explained. The study is based on an existing camera fault injection software that injects faults into the cameras of a working robot and collects the normal and faulty images recorded during this injection. The database obtained in the study is a source for the detection of anomalies that may occur in robotic systems. Within the scope of this study, a database of 10000 images consisting of 5000 normal and 5000 faulty images was created. Faulty images were obtained by injecting seven different types of image faults, namely erosion, dilation, opening, closing, gradient, motionblur and partialloss, at different times while the robot was operating.

*K*eywords  robotics · robot operating system · image database · fault injection

## 1  Introduction

With the widespread use of industrial robots, the communication security of the sensors and robots used in these robots with each other has become important. The data density that started to occur with the rapid development of the mentioned robot technologies and the healthy operation of this density has turned into a problem that forces the organizations that design and use these systems to take quicker decisions instantly. One way to operate and transfer this data traffic more quickly, efficiently and accurately is to detect abnormal events, instantaneous changes or deviations in this traffic at the same speed. To achieve this capture and to ensure the security of the systems, it is very critical to implement an artificial intelligence-based detection system. In this regard, many database studies lead to artificial intelligence studies that can be used to detect abnormal situations [1, 2, 3].

Detection of abnormal conditions in systems is a general expression of studies that do not conform to expected conditions or outputs, or other situations that are not visible to an expert in a database. Such abnormal situations can be turned into exploits such as structural faults or cheats by malicious people [4]. In order to establish artificial intelligence structures that can catch and report such abnormal situations, databases created from the outputs of the aforementioned systems in normal and abnormal situations should be used. Such system anomalies can be seen in various areas. To give an example from the field of botany, seeing weak or diseased plants during the growth of cultivated plants can also be considered as an anomaly. It is possible to detect this by examining a database created from the images of these plants [5]. As another example, a database can be created to examine the changes seen in human actions and these changes. This database can be powered by an artificial intelligence and the resulting changes can be detected [6]. Many more examples like this and studies on these examples make it possible

[1] Research and Development Department, Inovasyon Muhendislik Ltd. Sti., Eskisehir, Turkey
[2] Software Engineering Department, Eskisehir Osmangazi University, Eskisehir, Turkey

to create a database, to detect anomalies, and to obtain sufficient data to solve developing problems.

This use of databases allows accessible comparisons to be made to improve existing algorithm structures and try new techniques. Recent developments, especially in fields such as artificial intelligence and deep learning, have taken place faster with the discovery and creation of such databases under development. These databases are expanded to include more images as the number of studies increases, and their use in different fields is becoming more common day by day.

Deep behavior learning is one of the research areas where databases have a critical importance. Studies in this field include identifying and learning the features of the same people in different databases [7], recognizing the movements of people in videos obtained with a camera, determining their typical movements, using object recognition [8, 9,10] and using autonomously moving robots. It included features such as recognizing human mobility in the environment [6].

With the spread of the Internet, databases consisting of billions of image data can be accessed free of charge [1, 11, 12, 13]. Such databases have been published openly for researchers to collect and classify a large number of different types of recorded images, and as essential resources for other scientific research. Torralba et al. In the study by [13], a large database was created by collecting about 80 million images.

The databases used in these studies; augmented reality [14, 15, 16], robotics [17], autonomous drone navigation [18], forensic techniques [19], botany [5], data classification [25, 26], traffic control. [27], location detection [21], editing and correction of distorted images [3]. Taking these studies as an example, a database that can be used in the field of robotics was created in this study. The created database is designed to be a source for artificial intelligence studies that will enable the detection of image faults that may occur in robot cameras.

## 2    Creation of Camera Fault Injection Dataset

The camera fault injection database collected in this study is a database containing image distortions that may occur as a result of various malfunctions that a robot camera may encounter, and faulty pictures recorded as a result of anomalies. Problems that may be encountered can be thought of as the camera recording the corrupted images at that time and the system continuing to work on these corrupted images unnoticed. The created camera fault injection software and algorithms are designed to simulate such a situation. The basis of the software is to inject an fault into a working industrial robot system with a camera [32] and degrade the chassis images that the system should save with various image degradation techniques [24]. The database consists of normal images recorded by the working robot camera and degraded images after fault injection (Figure 1).

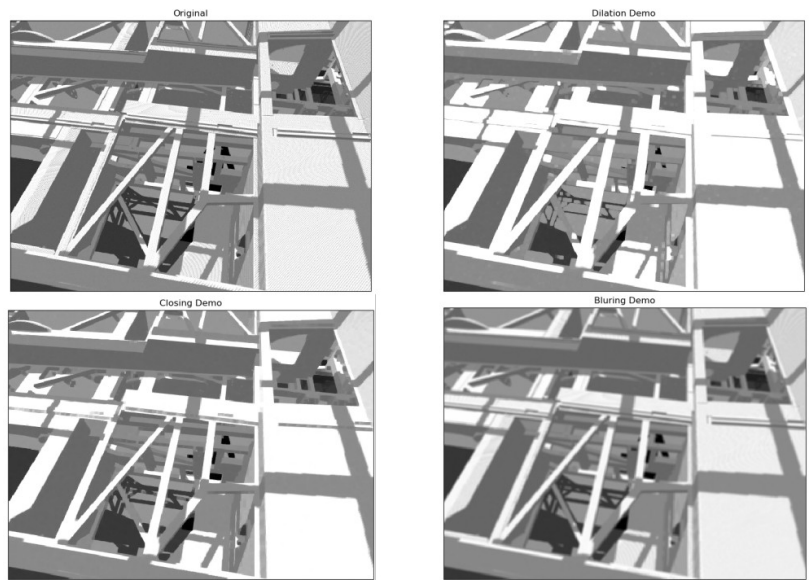

Figure 1: Camera fault injection outcome examples

## 3    Camera Fault Injection Dataset Contents

This study covers the camera sensors of the ROKOS robot control system working on the SRVT ecosystem and the VALU3S project "Evaluation Scenario- 4: Manipulation of Sensor Data" scenario [32]. This fault injection software sends faulty messages to the camera subjects in the ROS communication system via Python codes, and the robot cameras save these faulty images to its system while recording the incoming images.

### 3.1 Camera Fault Types

While creating a database from this system, 10000 images, 5000 normal and 5000 faulty images, were taken from the system. The fault types in which the generated 5000 defective images are revealed are listed below [22].

- **Dilation Method:** It is used to enlarge the highlighted parts of an image. There are 476 dilation faulty images in the database (Figure 2).

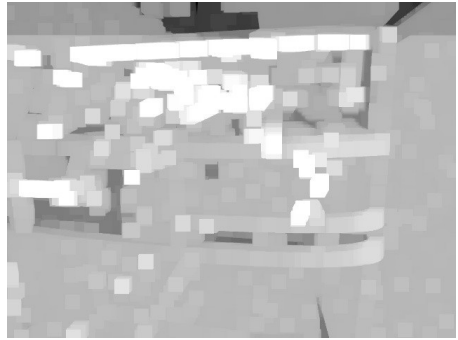

*Figure 2: Dilation Method example*

- **Erosion Method:** It is used to reduce the highlighted part of an image. There are 637 erosion faulty images in the database (Figure 3).

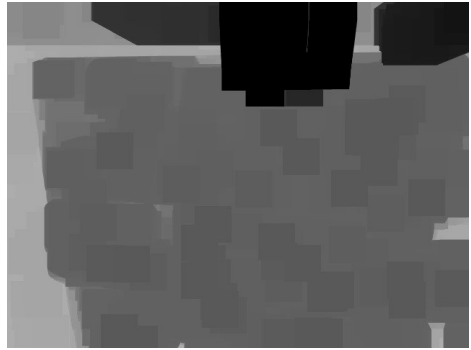

*Figure 3: Erosion Method example*

- **Open Method:** It is created by first etching and then spreading an image. There are 640 opening faulty images in the database (Figure 4).

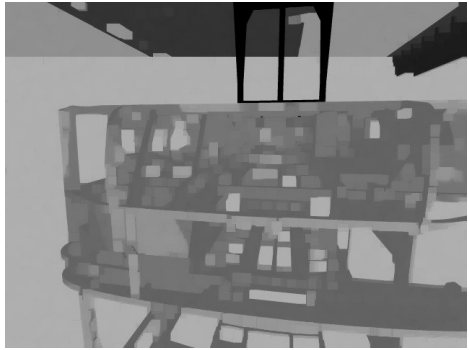
*Figure 4: Open Method example*

- **Close Method:** It is created by first spreading and then etching an image. There are 841 closing faulty images in the database (Figure 5).

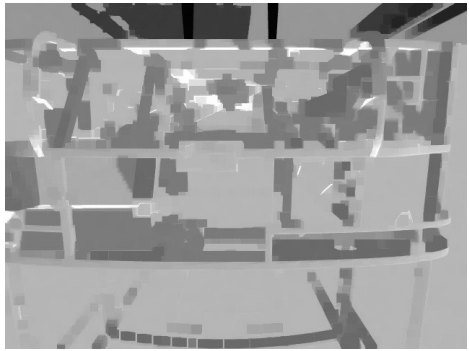
*Figure 5: Close Method example*

- **Gradient Method:** It is created by subtracting the etched version of an image from the smeared version. There are 632 gradient faulty images in the database (Figure 6).

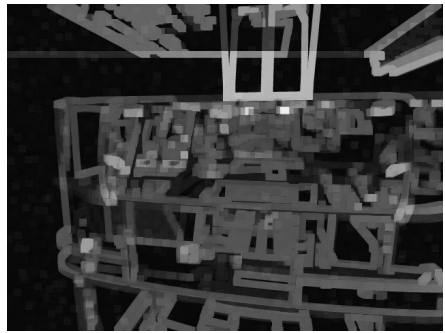
*Figure 6: Gradient Method example*

- **Motion-blur Method:** It is created by providing blur in an image. There are 687 motion-blur faulty images in the database (Figure 7).

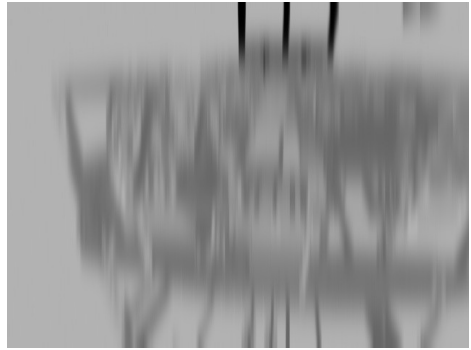

*Figure 7: Motion-blur Method example*

- **Partialloss Method:** Created by destroying a specified portion of an image. There are three different partialloss types: horizontal, vertical and odd-even. There are 1087 images of these three different types of partialloss fault in the database (Figure 8).

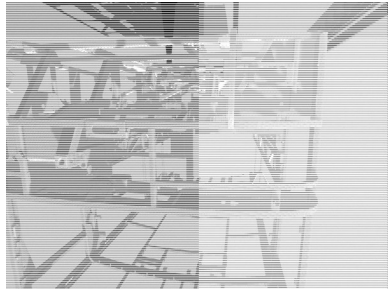            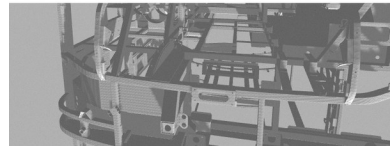

(a)                                                          (b)

*Figure 8: Partialloss Method examples*

The database created is available to everyone at:

https://drive.google.com/drive/folders/1pIqEnIIKFN13Z4m38zEB-XYQp_qb0T6S?usp=sharing.

## 4    Conclusion

In this study, a database consisting of faulty images obtained from robotic simulation was created in order to create a source for artificial intelligence applications to be used for anomaly detection in industrial robot cameras. In this context, it is explained how to create a database of 10000 images, how to collect faulty images obtained by fault injection into a robot camera and normal images recorded by this system, how many images of which fault types are found and how this happens. Thanks to this database, it is foreseen to create a useful resource for the development of artificial intelligence software that can detect the deterioration that may develop in industrial robot cameras.

This study has also been prepared to be used in "VALU3S_WP1_Industrial-1: Manipulation of Sensor Data" and "VALU3S_WP1_Industrial-4: Anomaly Detection at Component and System Level" scenarios, which are VALU3S project evaluation scenarios. For the first scenario, faulty data acquisition methods can be used. For the fourth scenario, a ready-made dataset will be provided to be used in studies where anomaly detection will be carried out.